%% file: main.tex
\documentclass{article}

\usepackage{microtype}
\usepackage{graphicx}
\usepackage{booktabs} 
\usepackage{tabularx}

\usepackage{hyperref}

\usepackage[utf8]{inputenc}
\usepackage[T1]{fontenc}
\usepackage{url}
\usepackage{todonotes}
\usepackage{amsfonts}
\usepackage{nicefrac}
\usepackage{dsfont}
\usepackage{MnSymbol}
\usepackage{float}
\usepackage{tikz}
\usetikzlibrary{bayesnet}
\usepackage{pgfplots}
\pgfplotsset{compat=1.14}
  \tikzstyle{latent} = [circle, fill=white,draw=black,inner sep=1pt,
minimum size=30pt, font=\fontsize{15}{15}\selectfont, node distance=1]
\tikzstyle{plate caption} = [caption, node distance=0, inner sep=0pt, font=\fontsize{15}{15}\selectfont,
below left=5pt and 0pt of #1.south east] %
  \tikzstyle{semi} = [circle, fill=white,draw=black,inner sep=1pt,
minimum size=30pt, font=\fontsize{15}{15}\selectfont, node distance=1, path picture={\fill[gray!25] (path picture bounding box.south) rectangle (path picture bounding box.north west);}]
\tikzstyle{plate caption} = [caption, node distance=0, inner sep=0pt, font=\fontsize{15}{15}\selectfont,
below left=5pt and 0pt of #1.south east] %

\tikzstyle{second plate caption} = [caption, node distance=0, inner sep=0pt, font=\fontsize{15}{15}\selectfont, below left=5pt and 0pt of #1.south west] %

\renewcommand{\plate}[5][]{ %
  \node[wrap=#3] (#2-wrap) {}; %
  \node[plate caption=#2-wrap] (#2-caption) {#4}; %
  \node[second plate caption=#2-wrap] (#2-caption) {#5}; %
  \node[plate=(#2-wrap)(#2-caption), #1] (#2) {}; %
}

\newcommand{\ltkiz}{1.5cm}

\usepackage{graphicx}
\usepackage{svg}
\usepackage{todonotes}
\usepackage{amsmath,amsthm,mathtools}

\usepackage{mathtools}
\usepackage{amsmath,amsthm}
\usepackage{soul}
\usepackage{blindtext}
\usepackage{tcolorbox}

\DeclarePairedDelimiterX{\infdivx}[2]{(}{)}{%
  #1\;\delimsize|\delimsize|\;#2%
}
\newcommand{\kld}[2]{\ensuremath{D_{KL}\infdivx{#1}{#2}}}

\usepackage[accepted]{icml2019}

\icmltitlerunning{A joint model of unpaired data from scRNA-seq and spatial transcriptomics}

\begin{document}

\twocolumn[
\icmltitle{A joint model of unpaired data from scRNA-seq and spatial transcriptomics for imputing missing gene expression measurements}




\icmlsetsymbol{equal}{*}

\begin{icmlauthorlist}
\icmlauthor{Romain Lopez}{UCB,equal}
\icmlauthor{Achille Nazaret}{UCB,X,equal}
\icmlauthor{Maxime Langevin}{UCB,X,equal}
\icmlauthor{Jules Samaran}{UCB,M,equal}
\icmlauthor{Jeffrey Regier}{UCB,equal}
\icmlauthor{Michael I. Jordan}{UCB}
\icmlauthor{Nir Yosef}{UCB}
\end{icmlauthorlist}

\icmlaffiliation{UCB}{Department of Electrical Engineering and Computer Sciences, UC Berkeley, USA}
\icmlaffiliation{X}{Centre de Math\'ematiques Appliqu\'ees, \'Ecole polytechnique, Palaiseau, France}
\icmlaffiliation{M}{\'Ecole des Mines de Paris, France }

\icmlcorrespondingauthor{Romain Lopez}{romain\_lopez@berkeley.edu}
\icmlcorrespondingauthor{Nir Yosef}{niryosef@berkeley.edu}

\icmlkeywords{scRNA-seq, smFISH, Imputation, Deep Generative Models}

\vskip 0.3in
]



\begin{NoHyper}
\printAffiliationsAndNotice{}  
\end{NoHyper}

\input{0_opening}

\input{1_generative_model}
\input{2_posterior_inference}

\input{3_performance_benchmarks}

\newpage

\bibliographystyle{icml2019}
 \bibliography{biblio}

\onecolumn

\end{document}

%% file: 0_opening.tex
Spatial studies of transcriptomes provide biologists with gene expression maps of heterogeneous and complex tissues such as the mouse nervous system~\cite{Zeisel2018}. A wide array of experimental protocols exist based on either single-molecule fluorescence in-situ hybridisation (smFISH)~\cite{Shah2016,OsmFISH}, a combination of imaging and sequencing (starMAP)~\cite{StarMAP}, or high-resolution RNA sequencing in-situ (e.g., slide-seq)~\cite{Rodriques2019}. The first two techniques have a high degree of sensitivity~\cite{OsmFISH} but suffer from the need of \emph{a priori} selecting a small fraction of genes to be quantified out of the entire transcriptome (varying from fifty for smFISH to a few hundreds for starMAP). The third technique is not limited to a predetermined subset of genes and in principle can capture any gene (out of thousands in the transcriptome). However, the number of genes efficiently captured in slide-seq measurements is substantially lower than what is obtained with standard (i.e., non-spatial) single-cell RNA sequencing (scRNA-seq), which is also more prevalent and easier to implement~\cite{Klein2015,Dropseq}.

As a consequence, an important research problem is to integrate spatial transcriptomics data with scRNA-seq measurements, in order to either transfer cell-type annotation from scRNA-seq to spatial datapoints~\cite{LIGER, Zhu2018}, impute gene missing in the spatial assay~\cite{SEURAT3} or to approximate the physical location in a tissue of cells measured with standard scRNA-seq~\cite{Satija2015}. State-of-the-art methods focus on embedding both spatial and standard datasets into a latent space--- using matrix factorization techniques (Liger and Seurat Anchors)~\cite{LIGER, SEURAT3}--- which is then corrected via mutual nearest neighbours~\cite{MNN} or quantile normalization. However, because (i) the embedding step relies on a linear model of the data though there is no basis for supposing linearity and (ii) the alignment is executed ad hoc, such methods will have a higher chance of overlaying samples who show little biological resemblance. In this manuscript, we focus on the problem of imputing missing genes in spatial transcriptomics data based on (unpaired) standard scRNA-seq data from the same biological tissue. Notably, our problem is related to domain adaptation methods (CORAL)~\cite{sun2016return} as well as unpaired image-to-image translation~\cite{zhu2017unpaired}, which has been applied to genomic data (MAGAN)~\cite{amodio2018magan}.

Since a certain fraction of the features (i.e., genes) is present in both datasets, we propose building up on recent advances in the field of domain adaptation~\cite{Ganin2017} and introduce \textit{gene imputation with Variational Inference} (gimVI), a deep generative model for integrating spatial transcriptomics data and scRNA-seq data which can be used to impute missing genes. gimVI is based on scVI~\cite{scVI}, with a different architecture as well as an alternative choice of conditional distributions to better take into account the technology-specific covariate shift. 

After describing our generative model (Section 1) and an inference procedure for it (Section 2), we compare gimVI to alternative methods on real datasets (Section 3). Our source code, based on PyTorch, is publicly available at \texttt{https://github.com/YosefLab/scVI}.

%% file: 1_generative_model.tex
\vspace{-0.2cm}
\section{The gimVI probabilistic model}
Figure~\ref{bayesnet} represents the probabilistic model graphically. 
\vspace{-0.3cm}
\paragraph{Shared biology}
For each cell $n$, binary variable $s_n$ specifies whether it was captured by the scRNA-seq or the spatial experimental protocol (such a framework can be extended to handle multiple datasets). Conditioned on nuisance variable $s_n$, we treat each cell as an independent replicate from the following generative process. Latent variable
\vspace{-0.1cm}
\begin{align}
z_n \sim \mathcal N(0, I_{d})
\vspace{-0.2cm}
\end{align}
is a low-dimensional random vector describing cell $n$. Such a latent variable is commonly interpreted as cell type or cell identity and can be used for downstream analysis such as clustering or visualization. Let $\mathcal{G}$ be the set of genes captured by the scRNA-seq experiment. We assume $\mathcal{G}$ is a superset of $\mathcal{G}'$, the set of genes captured by the spatial experiment. 
Let $f^\eta$ be a neural network with parameters $\eta$ and a softmax non-linearity on the last layer. Variable
\begin{align}
\rho_{n} = f^\eta(z_n, s_n)
\end{align}
is a value on the probability simplex. It can be interpreted as the expected normalized frequencies of each gene $g$ of a individual cell. This intermediate value has a biological interpretation and can be useful for downstream analysis such as imputation and detecting differentially expressed genes~\cite{scVI}. 
In this manuscript we focus on imputation, leaving hypothesis testing across modalities for future research.

\paragraph{scRNA-seq measurements}
Let $\mu_{n}, \sigma_{n} \in \mathbb{R}_+^2$ be the empirical mean and variance of the log-library size, fixed before inference. Latent variable
\vspace{-0.1cm}
\begin{align}
    \ell_{n} \sim \mathrm{LogNormal}(\mu_{n}, \sigma^{2}_{n})
\end{align}
represents a scalar nuisance factor for the scRNA-seq cells which corresponds to discrepancies of sequencing depth and capture efficiency. Let $f^\nu$ be a neural network with parameters $\nu$. Let $\theta \in \mathbb{R}^{|\mathcal{G}|}_{+}$ denote a vector of gene-specific inverse dispersion parameters, estimated via variational Bayesian inference. For each gene $g \in \mathcal{G}$, we treat each observed gene expression level $x_{ng}$ as independent conditionally on $\{\ell_n, z_n, s_n\}$ and sampled from an overdispersed count conditional distribution:
\begin{align}
x_{ng} &\sim \left\{\begin{array}{lr}
        \mathrm{ZINB}(\ell_n\rho_{ng}, \theta_g, f_g^\nu(z_n, s_n)), \\
        \text{~~~~~or} \\
        \mathrm{NB}(\ell_n\rho_{ng}, \theta_g),
        \end{array}\right.
\end{align}
which can be dataset specific~\cite{Svensson582064}. In the following experiments, we will use a zero-inflated negative binomial (ZINB) conditional distribution.

\paragraph{Spatial measurements} For this part of the model, only the genes in $\mathcal{G}'$ are captured. We therefore re-normalize our expected frequencies as
\begin{align}
   \forall g' \in \mathcal{G}', \rho'_{ng'} = \frac{\rho_{ng'}}{\sum_{g \in \mathcal{G}}\rho_{ng}}.
\end{align}
As we expect to see little to no technical variation because of size effects in spatial data (especially imaging~based technologies), we use observed random variable $\ell'_n$ to represent the number of transcripts in a given cell, which originated from the genes included in the assay. Let $\theta' \in \mathbb{R}^{|\mathcal{G}|'}_{+}$ denote a vector of gene-specific inverse dispersion parameters, estimated as before. For each gene $g' \in \mathcal{G}'$, we also treat each observed gene expression level $x'_{ng'}$ as independent conditionally on $\{\ell_n, z_n, s_n\}$. However, we model $x'_{ng'}$ as 
\begin{align}
x'_{ng'} &\sim \left\{\begin{array}{lr}
        \mathrm{Poisson}(\ell'_{n}\rho_{ng'}'), & \text{for smFISH} \\
        \mathrm{NB}(\ell'_{n}\rho_{ng'}', \theta'_{g'}), & \text{for starMAP}.
        \end{array}\right.
\end{align}
Such a choice of distribution is method specific and depends on the properties of the protocol. Notably, osmFISH~\cite{OsmFISH} is reported to have a nearly perfect sensitivity (i.e., capturing molecules originating from the genes included in the assay), which allows us to model it with a flat mean-variance relationship distribution such as Poisson. 
\input{bayes-net.tex}

%% file: bayes-net.tex
\begin{figure}[ht]
\centering
\scalebox{0.5}{
\begin{tikzpicture}[thick]
  \node[obs] (x) {${x}_{ng}$} ; %
  \node[obs, right = 2 * \ltkiz of x] (y) {$x'_{ng}$} ;
  \node[latent, above= 0.8 * \ltkiz of x, xshift=\ltkiz]  (z) {${z}_{n}$};
  \node[obs, above=0.8 * \ltkiz of z, xshift=0] (s) {$s_n$};
  \node[latent, above=\ltkiz of x, xshift=-\ltkiz] (l) {$\ell_n$};
  \node[obs, above= \ltkiz of y, xshift=0.5 * \ltkiz] (l_f) {$\ell'_n$};  
   \plate[inner sep=0.40cm, xshift=0.2cm, yshift=0.cm] {plate1} {(y)} {$G'$} {\color{blue}{FISH}}; %
    \plate[inner sep=0.40cm, xshift=0.2cm, yshift=0.cm] {plate2} {(x)} {$G$} {\color{blue}{RNA}}; %
   \plate[inner sep=0.40cm, xshift=0.3cm, yshift=0.cm] {plate3} {(z) (y) (x) (s) (l_f) (plate1) (plate2)} {N} {\color{blue}{~Cells}}; %
    \edge {s} {x};
    \edge {s} {y};
    \edge {l} {x};
    \edge {z} {x} ; %
    \edge {z} {y};
    \edge{l_f} {y};
\end{tikzpicture}
}
\caption{The proposed graphical model. Shaded vertices represent observed random variables. Empty vertices represent latent random variables. Edges signify conditional dependency. Rectangles (``plates'') represent independent replication.}
\vspace{-0.4cm}
\label{bayesnet}
\end{figure}
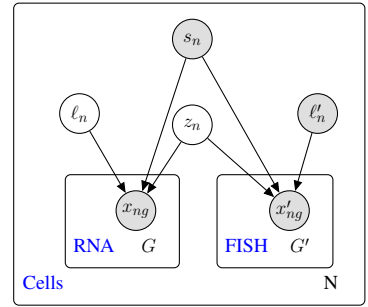

%% file: 2_posterior_inference.tex
\section{Posterior inference}

Bayesian inference aims at optimizing the data likelihood under the given model. Let $\Theta = \{\eta, \nu, \theta, \theta'\}$ denote the parameters of our generative model. The likelihood of the observed data decomposes as
\begin{align}
    \sum_{\substack{n=1 \\ s_n = 0}}^N\log p_\Theta\left(x_n \mid s_n\right) + \sum_{\substack{n=1 \\ s_n = 1}}^N\log p_\Theta\left(x'_n \mid s_n, \ell'_n\right).
\end{align}
As exact posterior inference is intractable, we use instead variational inference~\cite{AEVB} to approximate the posterior distributions. Our variational distributions $q_\phi(z, \ell \mid x, s) = q_\phi(z \mid x, s)q_\phi(\ell \mid x, s)$ and $q_\psi(z \mid x', s)$ are Gaussian with diagonal covariance matrices, parameterized by neural networks. The last layer is shared between the encoder networks for $q_\phi(z \mid x, s)$ and $q_\psi(z \mid x', s)$, which helps learning a joint latent space for the multiple modalities. The modality-specific variational lower bounds are 
\begin{align}
\begin{split}
        \log p_\Theta(x \mid s) &\geq \mathbb{E}_{q_\phi(z, \ell \mid x, s)}\log p_\Theta(x \mid z, s, \ell) \\ & - \kld{q_\phi(\ell \mid x, s)}{p(\ell)} \\ & -
    \kld{q_\phi(z \mid x, s)}{p(z)},
\end{split}
\end{align} 
\begin{align}
\begin{split}
        \log p_\Theta(x' \mid \ell', s) &\geq \mathbb{E}_{q_\psi(z \mid x', s)}\log p_\Theta(x' \mid z, s, \ell')  \\ & - \kld{q_\psi(z \mid x', s)}{p(z)}.
\end{split}
\end{align} 
We maximize these lower bounds using the stochastic backpropagation algorithm. 

\paragraph{Missing gene imputation} Let us derive a condition for which we can accurately impute a given gene $g \in \mathcal{G} - \mathcal{G}'$ for a given spatial measurement $x'$. To this end, we would like to sample $z$ from the posterior distribution $p_\Theta(z \mid x', s = 1)$ and use neural network $f_g^\eta$ to impute the counterfactual gene expression $x^*_g(z)$. Let $p_\text{spa}(z)$ (resp. $p_\text{seq}(z)$) denote the aggregated posterior for the spatial data (resp. scRNA-seq data). From domain adaptation theory~\cite{NIPS2007_3212}, for a given bounded loss function $\mathcal{L}$,  there exists a constant $\kappa$ such that the following bound holds
\begin{align}
\begin{split}
\label{domain_adapt}
\mathbb{E}_{p_\text{spa}(z)}\mathcal{L}\left(f_g^\eta(z), x^*_g(z)\right)\leq & \mathbb{E}_{p_{\text{seq}}(z)}\mathcal{L}\left(f_g^\eta(z), x^*_g(z)\right) \\ & +  \kappa d_\mathcal{H}(p_{\text{spa}}(z), p_{\text{seq}}(z)),
\end{split}
\end{align} 
where $d_\mathcal{H}$ designates the $\mathcal{H}$-divergence between the two latent spaces, and can be approximated by the loss of an adversarial classifier~\cite{Ganin2017}. This result explains that imputing a missing gene is possible whenever (i) such a gene is well fitted in the scRNA-seq data model and (ii) when the samples mix well in latent space.

%% file: 3_performance_benchmarks.tex
\section{Performance benchmarks}
To assess the performance of gimVI, we will present the statistical trade-offs for integrating cells into a joint latent space (Section 3.1) and then benchmark our method for imputing missing genes (Section 3.2). Throughout, we compare gimVI (for different values of $\kappa$) with vanilla scVI, Liger and Seurat. Only gimVI is able to make use of the held-out genes in $\mathcal{G}$ when learning the latent space --- a key advantage relative to other methods which focus on the gene set $\mathcal{G}'$ and only uses the complementary genes for downstream analyses like imputation. For benchmarking purposes, we run all the methods with $d = 10$ (dimension of the latent space).

We apply gimVI on two pairs of real datasets. First, we use a scRNA-seq dataset of 3,005 mouse somatosensory cortex cells~\cite{Zeisel2015} and a osmFISH dataset of 4,462 cells and 33 genes from the same tissue~\cite{OsmFISH} (referred to as mSMS). Second, we use a scRNA-seq dataset of 71,639 mouse pre-frontal cortex cells~\cite{Dropseq} and a starMAP dataset of 3,704 cells and 166 genes from the same tissue~\cite{StarMAP} (referred to as mPFC). For each pair of datasets, we hold out 20\% of the genes in $\mathcal{G'}$ and define $\mathcal{G}$ as $\mathcal{G'}$  with the addition of the held-out genes. In future work, we will investigate the model's robustness for wider ranges of ratio $\nicefrac{\mathcal{G'}}{\mathcal{G}}$. Since Liger throws a runtime error (probably a memory issue) on a Intel i7-4500U addressing 8GB of RAM while running on the mPFC pair, we randomly subsampled the corresponding scRNA-seq dataset to 15,000 cells for benchmarking purposes. We ran scVI and gimVI on a NVIDIA Tesla K80 GPU. For all the algorithms, fitting the data took less than a few minutes. 

\subsection{Integrating cells into a joint latent space}

\begin{figure}[ht]
    \centering
    \includegraphics[width=\columnwidth]{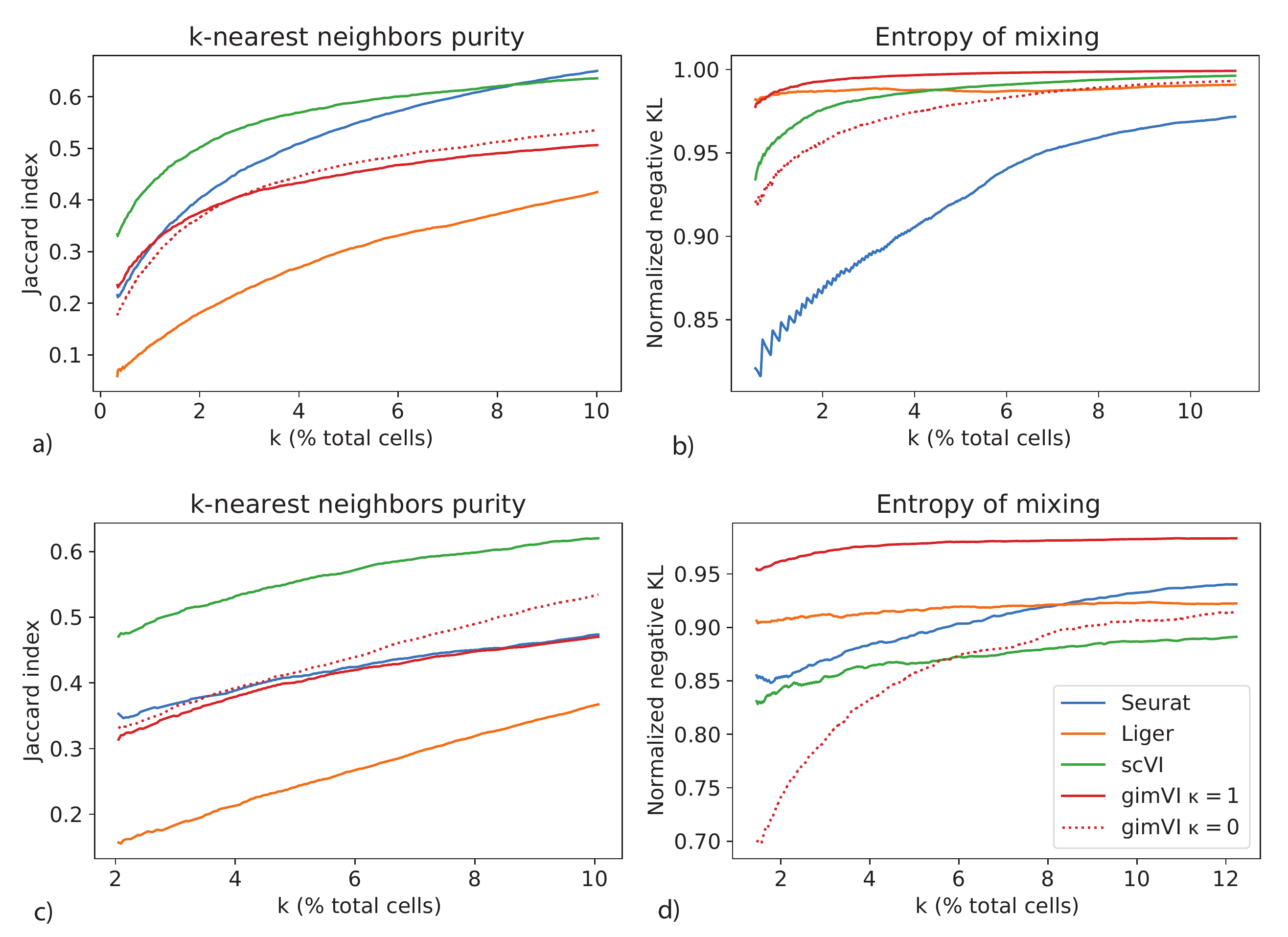}
    \vspace{-0.6cm}
    \caption{Integration metrics on the mPFC cells (a), (b) and the mSMS cortex cells (c), (d). 
    }
    \vspace{-0.5cm}
    \label{fig:harmo}
\end{figure}

We assess our model's ability to integrate information from two drastically different datasets. 
First we compute the entropy of mixing~\cite{MNN} to quantify how well the algorithms integrate the datasets in the latent space. As the pairs of datasets are unbalanced, we use the negative KL divergence between the $k$-nearest neighbors ($k$-NN) local $s_n$ distribution and the global $s_n$ distribution, which generalizes the entropy of mixing for that setting. Because this metric can be easily maximized by an algorithm that would ignore the input, we propose to use the $k$-NN purity~\cite{Xu2019} to evaluate whether an algorithm would return a similar latent space either by integrating the data or on individual datasets. For Seurat (resp. Liger, gimVI), we use PCA (resp. NMF, scVI) as the corresponding method for individual datasets and compute a Jaccard index to measure the overlap of the $k$-NN graphs. As such metrics could depend on the size of the neigborhood, we report these over a wide range of values for $k$ in Figure~\ref{fig:harmo}.

Across the two pairs of datasets, Liger has the lowest $k$-NN purity, suggesting that it may more significantly overlay sample of little biological resemblance compared to the other methods. Furthermore, gimVI has lower $k$-NN purity than scVI, which is expected since gimVI also uses its latent variables to impute genes which are unobserved in scVI. This behavior is exacerbated for higher values of $\kappa$ (see the statistical trade-off described in Eq.~\ref{domain_adapt}). In terms of entropy of mixing, gimVI with $\kappa = 1$ outperforms all methods. Notably, scVI has the second best median entropy of mixing for the mPFC cells, but the worse for the mSMS cells, which may be due to the inadequacy of the ZINB distribution to model osmFISH data. Overall, these results suggest that gimVI reaches the best trade-off (adjustable with $\kappa$) between merging the datasets and keeping intact the biological information.

\subsection{Imputing missing genes}
Since certain genes are held-out for benchmarking purposes, we propose to impute them from the unpaired scRNA-seq data and use the held-out information as groundtruth. As Liger,  and scVI do not explicitly model the unseen genes, we impute their measurements using a $k$-NN regressor trained on scRNA-seq data in the joint latent space--- we selected $k = 5\%$ number of cells. For Seurat, we used Seurat Anchors' ad hoc method for imputation. Furthermore, because spatial measurements and scRNA-seq have sensibly different distributions and gene-specific biases, we expect our imputed values to carry these biases from scRNA-seq. In particular, we choose to evaluate our model using the Spearman correlation $\rho$ over cells for each gene instead of the mean-square error, because the correct range of imputation may be unidentifiable without more information about the protocol-specific capture efficiencies. For this task, we also ran CORAL and MAGAN. However, MAGAN returned uniformly random values, despite our efforts to train the model (unshown). We report our results in Table~\ref{tab:cortex}.
\begin{table}[ht]
\small
    \centering
    \vspace{-0.15cm}
\begin{tabular}{l|cc|cc}
          & \multicolumn{2}{c}{mSMS}             & \multicolumn{2}{c}{mPFC}             \\
\textbf{} & $\widetilde{\rho}$ & $\widetilde{\delta_{\rho}}$                   & $\widetilde{\rho}$ & $\widetilde{\delta_{\rho}}$              \\ \hline
Seurat &  $0.15 $ & $-57\%$  &  $0.08 $ & $-55\%$  \\
Liger &  $0.22 $ & $-28\%$  &  $0.09 $ & $-55\%$  \\
scVI &  $0.20 $ & $-36\%$  &  $0.06 $ & $-65\%$  \\
CORAL &  $0.18 $ & $-38\%$  &  $0.17 $ & $-15\%$ \\
gimVI $\kappa = 1$ &  \textbf{0.30} & $-12\%$  &  \textbf{0.22}  & $-3\%$  \\
gimVI $\kappa = 0$ &  \textbf{0.33} & \_  &  \textbf{0.22}  & \_  \\ \hline
gimVI $\kappa = \kappa^*$ &  \textbf{0.37} & $+23\%$  &  \textbf{0.22}  & $+3\%$  \\
\end{tabular}
    \caption{Imputation results. For each algorithm and datasets, we report the median $\widetilde{\rho}$ over Spearman correlations $\rho$ and the median $\widetilde{\delta_{\rho}}$ over per gene relative change $\delta_{\rho}$ with respect to gimVI.}
    \label{tab:cortex}
    \vspace{-0.3cm}
\end{table}

First, we observe that the median of $\rho$ values are relatively low for all the algorithms. This can be explained by the fact that $\mathcal{G}'$ is a random subset of all the spatially transcribed genes in our experiment: thus, some of the held-out genes in $\mathcal{G}$ may have small correlation with the genes in $\mathcal{G}'$, making the imputation task significantly harder. To analyze the correlation distribution closely, we therefore report the median over genes of the relative change $\delta_\rho$ on $\rho$ with respect to gimVI. For both datasets, all versions of gimVI provides a significant improvement over state-of-the-art methods. For example, in the mSMS dataset, gimVI $\kappa = 0$ has more than 38 \% relative improvement on $\rho$ for more than half of the imputed genes compared to CORAL. Second, on the mSMS dataset $\kappa =1$ in gimVI decreases the imputation performance compared to $\kappa = 0$. However, it is possible to find $\kappa^* \in (0, 1)$ such that imputation is significantly improved over both $\kappa =0$ and $\kappa =1$ (which is reflective of Eq~\ref{domain_adapt}). This result is less significant in the mPFC dataset, which may be indicative of the similarity between starMAP and scRNA-seq datasets. In future work, we will propose a principled algorithmic procedure to choose this parameter. 

\begin{figure}[ht]
    \centering
    \vspace{-0.1cm}
    \includegraphics[width=\columnwidth]{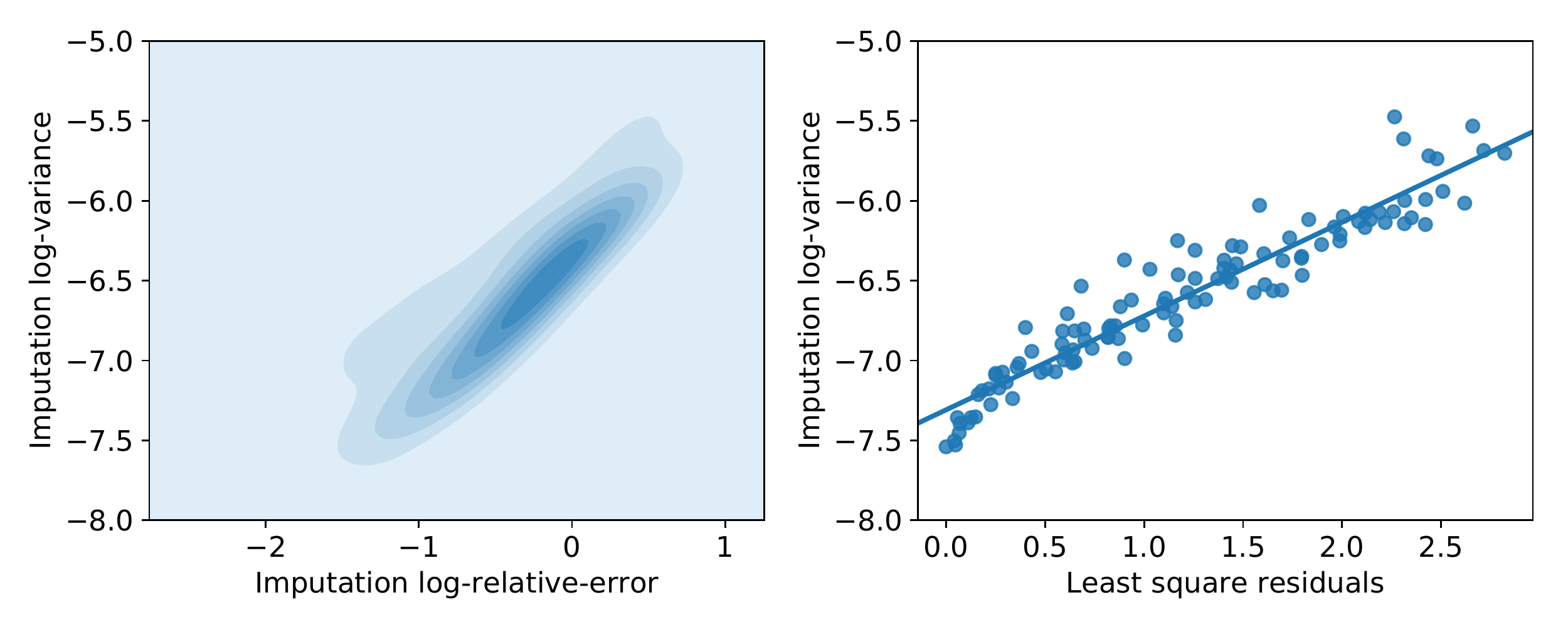}
    \vspace{-0.2cm}
    \caption{Imputation uncertainty of gimVI}
        \vspace{-0.2cm}
    \label{fig:uncertainty}
\end{figure}

Another fundamental difference between gimVI and other integration methods such as Liger and Seurat is that gimVI is a generative probabilistic method, able to carry uncertainty along with the imputed values. We can therefore derive the confidence of our model from the variance of its predictions. Comparatively, we computed for each gene the least-square residuals of a linear regression predicting hidden genes from observed genes on the scRNA-seq data only, to account for the ``predictability'' of the gene counts. Precisely, the variance on our model's prediction was computed by sampling fifty times from the variational posterior, and the linear regression was performed after normalization and log-transformation of the scRNA-seq data. We report our results for the mPFC cells in Figure~\ref{fig:uncertainty} and observe that the model is more uncertain for genes that are harder to predict. Similarly, gimVI has almost perfect predictions on values he is sure about (no variance). A next direction for research would be to detect which genes are worth imputing. 

\begin{figure}[ht]
    \centering
    \includegraphics[width=\columnwidth]{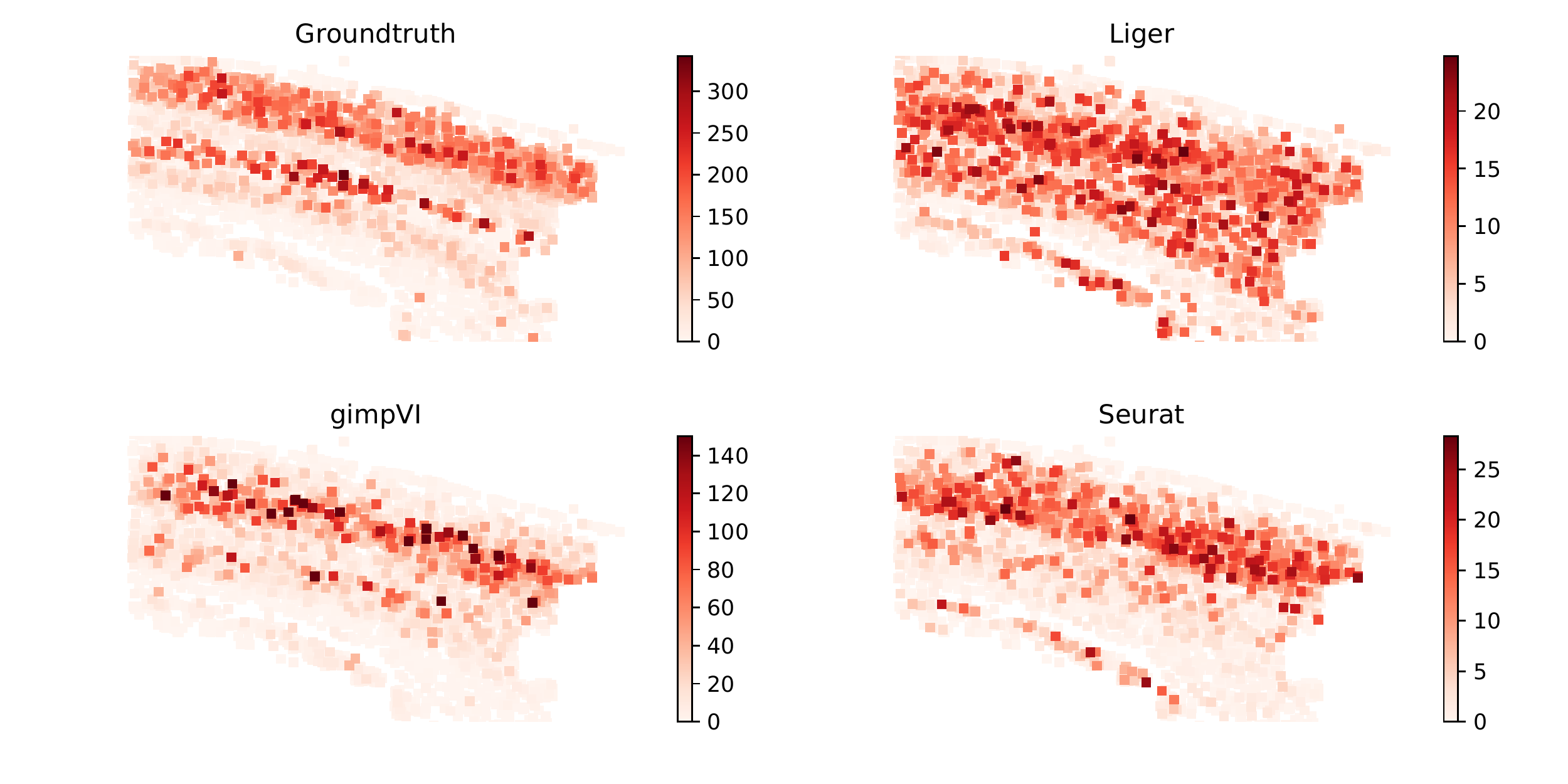}
    \vspace{-0.8cm}
    \caption{Discovering spatial motifs on imputed genes for the mSMS spatial dataset}
    \vspace{-0.2cm}
    \label{fig:spatial}
\end{figure}

A supplementary sanity check for the imputed values is to look at them in the context of discovering spatial patterns. In particular, we investigate held-out gene \emph{Lamp5}, a known marker gene for excitatory neurons and visualize them according to the cells' locations on Figure~\ref{fig:spatial} for the mSMS spatial dataset. Our result suggests that gimVI provides an imputation spatially more coherent than its competitors who guess high expression of this gene on wrong regions of the brain (e.g., layer 6). In future work, we will systematize this approach by extracting a curated list of positive and negative control and relying on hypothesis testing procedure to detect such patterns~\cite{Svensson2018}.
